# Commonsense Knowledge + BERT for Level 2 Reading Comprehension Ability Test

Yidan Hu[1], Gongqi Lin[2], Yuan Miao[2], Chunyan Miao[1]

1 Nanyang Technological University, Nanyang Ave, Singapore 639798
2 Victoria University, PO Box 14428, Melbourne, VIC 8001, Australia

**Abstract**

Commonsense knowledge plays an important role when we read. The performance of BERT [1] on SQuAD [2] dataset shows that the accuracy of BERT can be better than human users. However, it doesn't mean that computers can surpass the human being in reading comprehension. CommonsenseQA [3] is a large-scale dataset which is designed based on commonsense knowledge. BERT only achieved an accuracy of 55.9% on it. The result shows that computers cannot apply commonsense knowledge like human beings to answer questions. Comprehension Ability Test (CAT) [4] divided the reading comprehension ability at four levels. We can achieve human like comprehension ability level by level. BERT has performed well at level 1 which does not require common knowledge. In this research, we propose a system which aims to allow computers to read articles and answer related questions with commonsense knowledge like a human being for CAT level 2. This system consists of three parts. Firstly, we built a commonsense knowledge graph; and then automatically constructed the commonsense knowledge question dataset according to it. Finally, BERT is combined with the commonsense knowledge to achieve the reading comprehension ability at CAT level 2. Experiments show that it can pass the CAT as long as the required common knowledge is included in the knowledge base.

## 1   Introduction

Human brain is like a huge knowledge base. Commonsense knowledge is a part of this huge knowledge base and consists of a large number of well-known facts or relationships. We can learn many facts from daily life including reading. For instance, the color of grass is normally green. These facts will be stored in our brain and help us comprehend easier when we read and answer questions.

With the development of technologies, we hope machines can answer reading comprehension questions like people. Researchers designed a lot of natural language processing (NLP) models and datasets for this purpose. Among these models and datasets, BERT is the most notable state-of-the-art language model and SQuAD is one of the most widely used large scale datasets. SQuAD is composed of 536 articles from Wikipedia, and the questions and correct answers are given by crowd workers.

The experiment results on SQuAD show that BERT can achieve more than 87% accuracy in answering questions, while humans can only reach 82.3%. From analysis, we believe that BERT does not have specific ability to handle common knowledge. Even though it can perform well on some datasets statistically, its ability to handle common knowledge is very weak. In experiments, we found some slight modification to questions can easily fool BERT but giving human readers little trouble. Our findings are consistent to other researchers' report. To verify whether the existing NLP models can truly reach the level of human reading comprehension. There is a lot of work done on the adversary dataset based on commonsense knowledge. CODAH proposal a dataset based on five types of commonsense knowledge. they trained BERT on this dataset and find that BERT cannot learn and apply the commonsense knowledge to answer questions. The performance is just slightly better than random answers. commonsenseQA is another large-scale commonsense knowledge dataset. It is built based on conceptNet [4]. The dataset was tested with many NLP models. The result is similar to that on CODAH [5]. Although these dataset and tests have verified that the existing models cannot handle questions requiring commonsense knowledge, there is not clear clue on how commonsense knowledge can be incorporated in the reading comprehension. One of the important reasons is that there are many different types of commonsense knowledge, at different difficult levels. Each type of commonsense knowledge requires different processing models. CAT [4] classified the comprehension ability into four levels, while BERT has failed badly at level 2. We will move one step forward to improve machine comprehension ability to level 2.

In this work, we present a system which consists of three parts to help computers learn commonsense knowledge and answer reading comprehension questions. The first part of this system is building a commonsense knowledge graph. Each triple in the commonsense knowledge graph contains two entities and a relation between the two entities. According to the relationship between the two entities in commonsense knowledge, we divide the commonsense knowledge graph into three categories, attributes, synonyms and definitions respectively. In the attribute category, the entities are extracted from SQuAD and conceptNet, then we established the relationship from the Wikipedia of two entities in Wikipedia. In the synonyms and definitions categories, the entities are built based on wordnet, and the relationship between the two entities is the "synonym_is" and the "definition_is" respectively. The second part of the system the commonsense knowledge dataset which is based on the commonsense knowledge graph and SQuAD dataset. In the cases that the two entity relationships are synonyms and definitions in the commonsense knowledge graph, we interchanged the two entities in the question to create new questions. When an entity-relationship belongs to attribute, we replaced the relationship and entity in the question at the same time. In the final part, we combined the commonsense knowledge dataset and fine-tuned BERT to answer question requiring commonsense knowledge.

To summarize, our contributions are:

- We built a commonsense knowledge graph (knowledge base) based on SQuAD, conceptNet and wordnet.

- We built a commonsense knowledge dataset based on the commonsense knowledge graph.
- We experimented BERT and BERT + commonsense knowledge to test how well they perform at reading CAT at level 2, confirmed that BERT failed badly at this level, while BERT + commonsense knowledge can achieve similar performance of what BERT achieved at CAT level 1.

## 2 Comprehension Ability Test

### 2.1 CAT level 1 and level 2

CAT level 1 is easier than level 2 for NLP models. In CAT level 1, the NLP models need to find the fact which can be find in the article and don't need to analysis or reason. At this level, BERT as the state-of-the-art NLP model can answer questions extremely well even surpass the human being. However, CAT level 2 should apply commonsense knowledge to answering questions. The study of CommonsenseQA shows that it is hard for many NLP models including BERT. Since CommonsenseQA consists of many questions created with commonsense knowledge, the performance of BERT is similar with random result. It will significantly limit the scope of use of BERT. When we need NLP models to read and answer questions, we cannot force every questioner to use the suitable terms or words. To address this problem, we implement the CAT level2 to help NLP models raading with commonsense knowledge.

### 2.2 Commonsense knowledge graph

Knowledge graph consists of triples. Each triple represents a relationship between two entities. Commonsense knowledge graph contains knowledge in the triples. We can represent a triple in commonsense knowledge graph as <*t,r,n*>. In this triple, t represents a commonsense subject; *n* represents a commonsense object; r represents the relationship between the commonsense subject *t* and the commonsense object *n*. We divide the relationship *r* into three categories: attribute, synonym and the definition respectively, based on the types of commonsense knowledge.

Attribute is objects' feature and property which is also a equivalent relation of two objects in one context. Object can have many attributes, and these attributes can help us distinguish it from other objects. For instance, the color of snow is white. In this sentence, snow is the commonsense subject and white is the commonsense object. Color is not only the relationship but also belong to the attribute of snow. It defines a conditional equivalent relationship. We can say snow and white are equivalent under the condition of color.

The commonsense knowledge in this definition can be constructed from scratch. But given that SQuAD has been widely applied, we started to build one for it. In order to construct a commonsense knowledge map with attributes as a relationship, we extracted the commonsense subjects from the SQuAD dataset. We put these commonsense subjects into a set T.

We then collect commonsense objects from conceptNet. ConceptNet is a freely available commonsense knowledge base and semantic network. It contains knowledge related to the world. These knowledge can help computers better answer questions. It consists of a number of nodes that represent concepts expressed in natural language words or phrases. We mainly applied the "related terms" in conceptNet in this work, but there are no essential difficulties to be extended to other relationships. We can apply the same method and collect all the related terms of the subjects (e.g. snow) in conceptNet.

Each commonsense subject can have several commonsense objects, which are distinguished by the relationships. The relationship of conceptNet is too general to make sufficient sense. For example, snow has a related term white. ConceptNet tries to reduce the number of relationships and thus lost the actual commonsense. Here the commons sense is that snow's color is white. We construct the commonsense relationship by extracting the corresponding wikis from Wikipedia. We compared the similarities of the words in the definition or the whole wiki of the commonsense subject and the commonsense object. For instance, the definition of white in the Wikipedia is, "White is the lightest color and is achromatic (having no hue). It is the color of fresh snow, chalk, and milk, and is the opposite of black." By extracting the relationships, we can get the relationship between snow and white, which is color.

Synonym is another type of equivalent relationship. Synonym means that two words have the same meaning and can be replaced with each other in a specific context. For example, given the question " Why did Henry buy a new laptop for programming?", we can change "laptop" to "notebook" (a synonym), resulting in the question, "Why did Henry buy a new notebook for programming?". There are not many universal synonyms. Notebook can also mean paper book we write notes. The context when notebook and laptop are synonyms is when we talk about computing devices. So we can see the relationship here is the context of computing devices, or (notebook, computing devices, laptop).

Definition refers to the interpretation of a word. For example, given the question "What happened to my hometown in the decade?", we would change "decade" to "a period of ten years" (a definition), resulting in the question, "What happened to my hometown in a period of ten years?" Many words have multiple meanings thus multiple definitions. The definition of a word will be a specific meaning according to the specific context. For example, apple can be a type of fruit, or an IT company. The relationship refers to the context when the corresponding definition holds. We collected synonyms and definitions types of commonsense from WordNet.

## 2.3 Dataset generation
In order to verify whether the NLP models can answer questions with commonsense knowledge, we need to construct a commonsense knowledge dataset. We build a commonsense knowledge dataset based on our commonsense knowledge graph. We extracted the questions which include the commonsense subjects in SQuAD and generated the relationship based on the commonsense knowledge graph. For example,
- given the sentence " The top of Mount Fuji is covered with snow.",

- and the question, "What does the top of Mount Fuji have?",
- the answer is "snow".

we can change this question into a new question because this question contains the commonsense subject 'snow'. From the commonsense knowledge graph, we can get the relationship between snow and other commonsense objects, such as color and temperature and so on. Then we use the relationships to modify the question. The new questions can be
- "What color does the top of Mount Fuji have?" or
- "What is the temperature on top of Mount Fuji?"

Currently we have manually built a dataset called S-ETCK which includes 100 questions, and We will automatically build a large dataset in the future. Questions in S-ETCK is created based on the question in SQuAD. Since answers to questions in SQuAD can be found almost in the article, Bert can answer it very well. To interfere with Bert, we replace the words in the original question with attributes, synonyms and definitions.

2.4 Algorithm verification

To verify whether machines can use commonsense to answer questions. We implemented our algorithm based on BERT. We firstly check if the words or context of the question is a relationship in the commonsense knowledge graph. If yes, we get the commonsense subject from the original article according compare the similarity between the question and the sentence in the article. We compare the commonsense subject in the commonsense subject list with all the words in the target sentence and get the word of highest similarity as commonsense subject. According to the known commonsense subject and relationship, the commonsense object can be extracted from the triple in commonsense knowledge graph to construct the answer. It is worth noting that our algorithm does not affect the accuracy of answers to normal questions which do not contain commonsense questions. It means that we will improve Bert's reading comprehension ability. Our algorithm always can handle problems even though the people use different words to express same thoughts. In order to achieve this goal, we will only need to verify the questions that BERT can't answer. These questions are all that requires commonsense knowledge to answer.

**Algorithm 1**

**Input:** Article a, Question q, Commonsense knowledge graph c
**Output:** Answer s
   Initial S=[], tmp=[]
   S= all possible answers of Bert output
   **For** i in s:
     If (c.t in i):
       tmp+= tripes of c.t
   **for** j in q:
     for t in tmp:
       if(j==t.r)
         **return** t.n

## 3 Experiment

### 3.1 BERT

We evaluated a pre-trained BERT-Base which is a state-of-the-art language model in a range of tasks. This model contains a 12-layer network, with 768 hidden units per layer, 12 self-attention heads and a total of 110M parameters. BERT implements bidirectional transform with a masked language model. masked language model randomly masks some of the tokens from the input, and the objective is to predict the original vocabulary id which was masked.

### 3.2 BERT Hyperparameter

We apply $BERT_{BASE}$ (uncased) as the basis for our experiments. The maximum total input sequence length after WordPiece tokenization is 384. The maximum number of tokens for the question is 64. The maximum length of an answer that can be generated is 30. The total number of training epochs to perform is 3.0. We set the training batch size as 4. We used a GPU with memory of 12G. We used Adam optimizer and set the learning rate to 5e-5. We trained 48,869 steps for fine-tuning the model based on SQuAD 2.0 dataset.

### 3.3 Evaluation Metrics and Model Selection

We used the same evaluation script from SQuAD. It uses Exact Match (EM) and F1 scores. EM and F1 scores ignore punctuations and articles. Recall reflects the ability of the classification model to identify positive samples. The higher the recall, the stronger the recognition ability of the model for positive samples. The precision reflects the ability of the model to distinguish negative samples. The higher the precision and the difference between the model and the negative samples are, the more powerful is the model. F1-score is a combination of the precision and the recall. The definition of F1 is shown in formula (1). The higher the F1-score is, the more robust the classification model is.

$$F1 = \frac{2TP}{2TP+FN+FP} = \frac{2 \cdot Precision \cdot Recall}{Precision+Recall} \quad (1)$$

The EM requires the answers to have exact string match with human-annotated answer spans. The final F1 score is the averaged F1 scores of individual answers, which is typically higher than EM. F1 is the major evaluation metric. Each individual F1 score is the harmonic mean of individual precision and recall computed based on the number of overlapped words between the predicted answer and human-annotated one. This metric measures the percentage of forecasts that exactly match any of the ground truth answers.

### 3.4 Result

Results for the above configurations are shown in Table 1. we can find that BERT has an outstanding performance on SQuAD, up to 71.862% F1 score. However, it also shows that BERT had a very poor performance on our commonsense knowledge dataset. The F1 score decreased to 48.557%. The SQuAD questions were changed to new questions by replacing the original words or phrases to their attributes, synonyms or definitions. Therefore, answering these questions requires commonsense knowledge. It gives human users little extra difficulties. Besides, the EM also dropped a lot from 71.862% to 38.613% on the S-ETCK dataset.

|  | EM | F1 |
|---|---|---|
| SQuAD | 71.826 | 75.487 |
| S-ETCK | 38.613 | 48.557 |

Table1: The EM and F1 of BERT on SQuAD and S-ETCK

|  | EM | F1 |
|---|---|---|
| BERT | XXX | XXX |
| BERT+Common Knowledge Graph Model | XXX | XXX |

Table2: The EM and F1 of BERT and BERT+Common Knowledge Graph Model on S-ETCK

The results of BERT and CAT on S-ETCK are presented in Table 2. BERT+Common Knowledge Graph Model should be similar on SQuAD and S-ETCK but takes a bit longer time beacuse it need to check commonsense knowledge.

**Conclusion**

We proposed a new system to help computers answer reading comprehension questions with commonsense knowledge. In this system, we built a commonsense knowledge graph including three equivalence types of commonsense relationship based on SQuAD, conceptNet, WordNet and Wikipedia. On this basis, we built a commonsense knowledge dataset. We trained an algorithm on our commonsense knowledge dataset to answer questions with commonsense knowledge like what human beings do. This algorithm is designed by combining BERT and commonsense knowledge graph.